\documentclass[10pt,twocolumn,letterpaper]{article}

\usepackage[pagenumbers]{cvpr} 

\usepackage[table]{xcolor}

\usepackage{bbm}

\definecolor{cvprblue}{rgb}{0.21,0.49,0.74}
\usepackage[pagebackref,breaklinks,colorlinks,citecolor=cvprblue]{hyperref}

\def\paperID{*****} 
\def\confName{CVPR}
\def\confYear{2024}

\title{Watch Your Step: Optimal Retrieval for Continual Learning at Scale}

\author{Truman Hickok\\
University of Texas at San Antonio\\
{\tt\small truman.hickok@utsa.edu}
\and
Dhireesha Kudithipudi\\
University of Texas at San Antonio\\
{\tt\small dk@utsa.edu}
}

\begin{document}
\maketitle
\begin{abstract}
In continual learning, a model learns incrementally over time while minimizing interference between old and new tasks. One of the most widely used approaches in continual learning is referred to as replay. Replay methods support interleaved learning by storing past experiences in a replay buffer. Although there are methods for selectively constructing the buffer and reprocessing its contents, there is limited exploration of the problem of selectively retrieving samples from the buffer. Current solutions have been tested in limited settings and, more importantly, in isolation. Existing work has also not explored the impact of duplicate replays on performance. In this work, we propose a framework for evaluating selective retrieval strategies, categorized by simple, independent class- and sample-selective primitives. We evaluated several combinations of existing strategies for selective retrieval and present their performances. Furthermore, we propose a set of strategies to prevent duplicate replays and explore whether new samples with low loss values can be learned without replay. In an effort to match our problem setting to a realistic continual learning pipeline, we restrict our experiments to a setting involving a large, pre-trained, open vocabulary object detection model, which is fully fine-tuned on a sequence of 15 datasets.

\end{abstract}
    
\section{Introduction}
\label{sec:intro}

The field of continual learning represents a cornerstone in the evolution of artificial intelligence, as continual learning algorithms open the door to agents that can efficiently self-improve in the face of failure and uncertainty, along with being able to adapt to changing requirements \cite{verwimp2023continual, hadsell2020embracing}. Fundamentally, continual learning algorithms seek to allow machine learning practitioners to sequentially train models on task-specific datasets following a broad, expensive pre-training phase, without sacrificing performance on pre-trained or downstream tasks. This setting differs from the current mainstream paradigm of either superficially adding new capabilities through in-context learning \cite{dong2023survey, balažević2023incontext} and retrieval augmentation \cite{lewis2021retrievalaugmented} or, much worse, appending new-task data to the pre-training dataset and retraining the model from scratch \cite{bousmalis2023robocat}. Therefore, continual learning algorithms uniquely allow new information to reconfigure the model's existing representations such that performance can be maximized on both novel \textit{and} existing tasks \cite{balaguer2024rag}.

Replay-based continual learning algorithms selectively store or generate inputs from previous tasks so that they can be mixed, or "interleaved", with data from new tasks \cite{hayes2021replay}. A key set of algorithms in this ecosystem is \textit{selective retrieval} algorithms, which retrieve samples from the replay buffer such that they are \textit{prototypical} and \textit{representative} of previous tasks \cite{harun2023grasp} or are likely to be \textit{interfered} with by samples from new tasks \cite{saxena2022learning, aljundi2019online}. However, up to this point, algorithms for selective retrieval have been evaluated in \textit{isolated} and \textit{limited} problem settings, which means that they have not been directly compared to each other and have been tested in settings with questionable applicability to modern continual learning tasks. Moreover, two basic properties of these algorithms have gone unexamined to this point: the manner in which duplicate replays should be avoided and whether each new sample should warrant replay regardless of the magnitude of its loss. 

In this paper, we expand on these works; we start with a broad question: \textit{what is the best algorithm for selective retrieval?} We begin answering this question by organizing and extending existing algorithmic \textit{primitives} for selective retrieval, which are the building blocks of the algorithms under study and are separated into \textit{class-selective} and \textit{sample-selective} primitives. Next, we specify existing \textit{and} novel combinations of primitives and compare their performance on a unique benchmark meant to match a modern, large scale continual learning setting (Figure~\ref{fig:CLaS}, bottom). We jointly execute this comparison with a comprehensive sweep of \textit{deduplication schedules}, which determine how long training takes without allowing duplicate replays. Interestingly, we observe that two of the simplest primitives, implemented as standalone solutions, perform better than all other algorithms, including a combination of these primitives themselves. As expected, deduplication is necessary for almost every algorithm, but only up to a certain shared point.

Motivated by recent work showing abrupt representation drift upon entering a new task \cite{sarfraz2023error, caccia2022new}, we then experiment with the idea of reducing the impact of replay when new samples' losses are below a threshold, along with reducing the number of replays in these cases. This procedure reflects the hypothesis that reducing the impact of replay for new samples with the smallest gradients can lead to extra plasticity for new tasks, without incurring more forgetting of previous tasks. Here, we find that applying this procedure rapidly leads to forgetting previous tasks, as the case where only \textit{3\%} of new samples are below the loss threshold leads to significant drops in performance.

Finally, we provide analyses regarding the distributions over samples and classes produced by our algorithms, as well as forgetting dynamics according to properties of our datasets. In summary, we:
\begin{enumerate}
    \item Compare algorithms for selective retrieval, demonstrating that certain algorithms outperform all others, including combinations of the algorithms themselves
    \item Compare different intervals over which no duplicate replays are allowed, demonstrating that duplicate replays should be prevented for each downstream \textit{dataset} 
    \item Show that only using replay for new samples with high losses quickly leads to an unacceptable drop in overall performance
    \item Analyze distributions produced by the best algorithms and discuss dataset-dependent forgetting dynamics
\end{enumerate}

\section{Related Work}

\textbf{Continual Learning:} The space of solutions in continual learning can be organized into three types of methods: regularization, model expansion, and replay \cite{hadsell2020embracing}. Regularization-based methods involve adding terms to the model's loss function to prevent the model from overwriting existing information, and can be executed in the space of weights or predictions \cite{smith2023closer, asadi2023prototypesample}. Model expansion involves adding new parameters to the network for each new task \cite{rusu2022progressive}. Replay-based methods involve sampling from a stored subset of data from previous tasks or from a generative model of previous tasks and interleaving data from new tasks with data from previous tasks \cite{buzzega2020dark, hayes2021replay}.

\begin{figure}[t]
  \centering
  \includegraphics[width=\columnwidth]{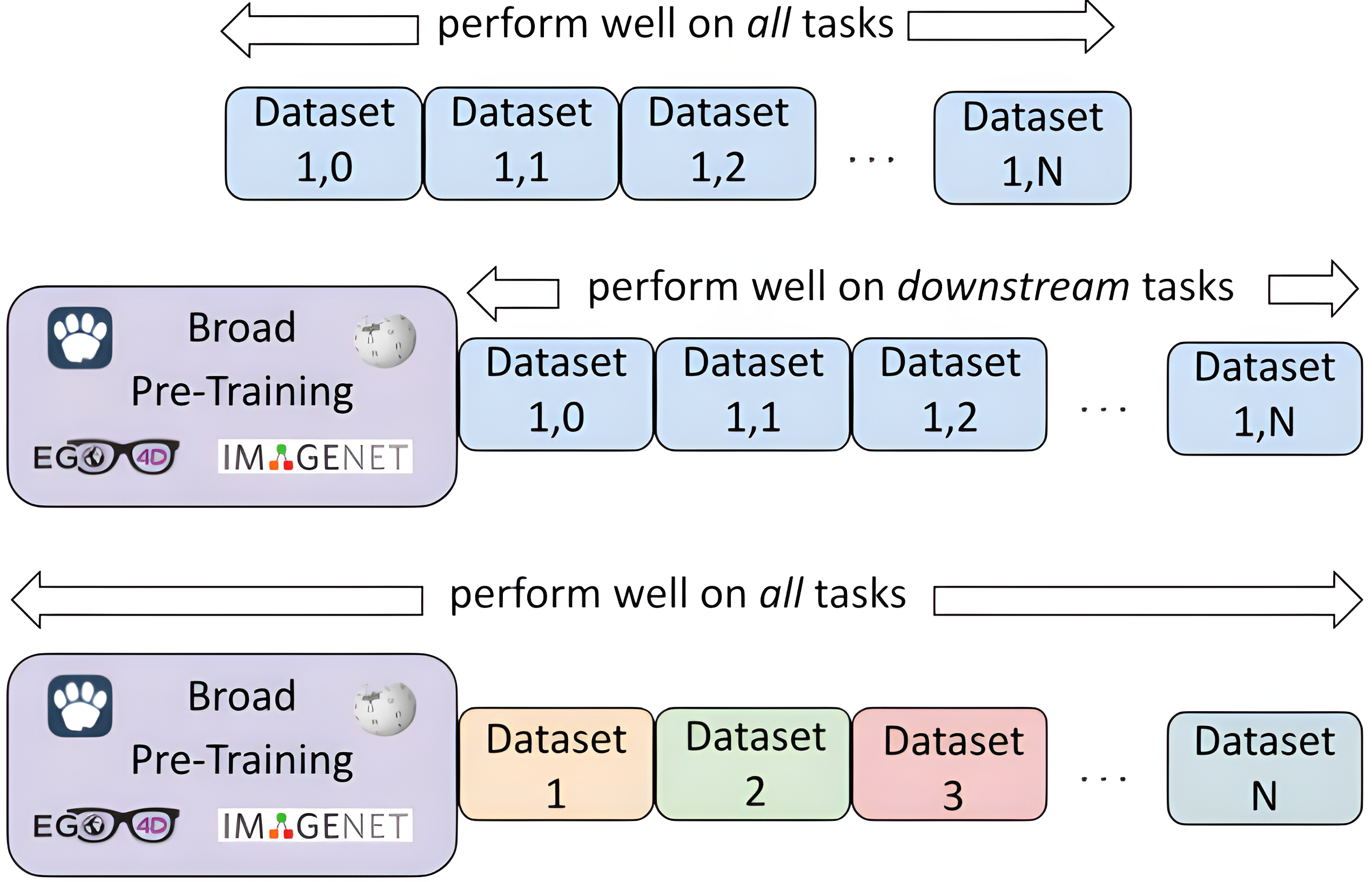}
  \caption{The problem settings of continual learning research. \textit{Top}: the original and most common setup, where dataset is divided into N balanced subsets and the model is trained sequentially on each subset. \textit{Middle}: another common setup, which is the same as above except the model is pre-trained; forgetting on pre-training tasks is \underline{ignored}. \textit{Bottom}: our setup, where a pre-trained model is sequentially trained on OOD \textit{datasets} and forgetting on pre-training tasks is \underline{minimized}.}
  \label{fig:CLaS}
\end{figure}

Almost all existing work in continual learning has been executed in the setting where a model is randomly initialized before being trained sequentially on balanced, independent subsets of a single dataset (Figure~\ref{fig:CLaS}, top) \cite{Delange_2021, vandeven2019scenarios}. On the other hand, a more recent line of work has considered the setting where a model is initialized after broad pre-training and then trained in the same fashion as the original setting (Figure~\ref{fig:CLaS}, middle) \cite{lee2022pretrained, panos2024session, janson2023simple, pmlr-v232-galashov23a}. This scenario is quite different from what modern machine learning pipelines require, as they need to continuously evaluate and preserve the extensive capabilities of the model during the learning process. As the field has progressed, several recent works address this issue \cite{ilharco2022patching, garg2023ticclip, lin2022clear} (Figure~\ref{fig:CLaS}, bottom).

\textbf{Replay Methods:} Replay algorithms can be organized into four distinct branches, which are not mutually exclusive. The first branch is concerned with how information from past tasks is represented. This class of methods varies along two axes: whether the data from previous tasks is \textit{stored} \cite{buzzega2020dark} or \textit{generated} \cite{shin2017continual, van2020brain} and whether \textit{inputs} \cite{buzzega2020dark} or \textit{activations} \cite{van2020brain, hayes2020remind} are being represented \cite{hayes2021replay}. The second branch is concerned with how replayed samples are \textit{reprocessed}, and includes regularization terms such as logit distillation \cite{buzzega2020dark} and gradient episodic memory (GEM) \cite{chaudhry2019efficient}. The third branch is concerned with selectively \textit{storing} information from previous tasks and encourages the samples in the replay buffer to be diverse \cite{tiwari2022gcr, aljundi2019gradient} and "easy" \cite{hurtado2023memory}. The fourth branch is concerned with selectively retrieving samples from the replay buffer and is the focus of this work.

\textbf{Selective Retrieval for Replay:} While the vast majority of work in continual learning has used balanced or fully random retrieval during replay \cite{buzzega2020dark, hayes2021replay}, some selective retrieval algorithms have recently been shown to outperform these baselines. Adversarial Shapley Experience Replay (ASER) \cite{shim2021online} proposes a scoring mechanism over samples in the replay buffer which balances terms representing how \textit{adversarial} each sample is with respect to the current batch and how \textit{representative} each sample is of other samples in the replay buffer. ASER was evaluated relative to MIR \cite{aljundi2019online} and random retrieval on Split-CIFAR and Split-Mini-ImageNet benchmarks in the traditional continual learning environment (Figure~\ref{fig:CLaS}, top). Similarity-Weighted Interleaved Learning (SWIL) \cite{saxena2022learning} proposes a weighted sampling distribution over \textit{classes} according to how similar each class's prototype is to samples in new batches. SWIL was only evaluated relative to balanced retrieval and was tested on CIFAR-10 and CIFAR-100 where 90\% of the classes are learned, then a \textit{ single new class} is learned with replay. "Gradually Select Less Prototypical" (GRASP) \cite{harun2023grasp} proposes a weighted sampling distribution over \textit{samples} according to how prototypical each sample is with respect to its corresponding class prototype. Although GRASP was evaluated in both a modern setting and relative to a slew of simple retrieval methods, it was not compared to SWIL or ASER. In addition to being tested in isolation, none of these methods considers the problem of deduplication or the prospect of loss-conditioned retrieval.

These three algorithms are the focus of this work and will be organized and extended (Section~\ref{sec:methods}) then compared (Section~\ref{sec:exps}) and analyzed (Section~\ref{sec:analyses}). 
\section{Problem Setting}

We consider the general continual learning setting where the learner is initialized as a pre-trained network and is then fine-tuned on a sequence of datasets (Figure~\ref{fig:CLaS}, bottom). Formally, after initializing parameters $\theta_{\text{pt}}$ using dataset $D_{\text{pt}}$, we sequentially fine-tune the network on $N$ datasets, which we refer to as $D_{\text{cl}}=\{D_{\text{cl,1}},D_{\text{cl,2}},...,D_{\text{cl,N}}\}$. Throughout this process, we aim to maintain (or improve) performance on all pre-training tasks while also improving performance on $D_{\text{cl}}$ as much as possible, which we accomplish by selectively retrieving and interleaving samples from the replay buffer, which is a subset of $D_{\text{pt}}$. We do not add samples from $D_{\text{cl}}$ to the buffer. When training without loss adaptivity, we append samples from the replay buffer to the current batch such that there is a 1:1 ratio between new samples and replay samples.
\section{Methods}\label{sec:methods}

In this section, we first formalize the principal objects of this paper: the primitives which constitute algorithms for selective retrieval during replay. We then describe our set of deduplication schedules, loss-thresholded replay, and our replay buffer selection algorithm. See Table~\ref{tab:algo} for an overview of all \textit{combinations} of primitives evaluated in our main experiments.

\begin{figure}[t]
  \centering 
  \includegraphics[width=\columnwidth]{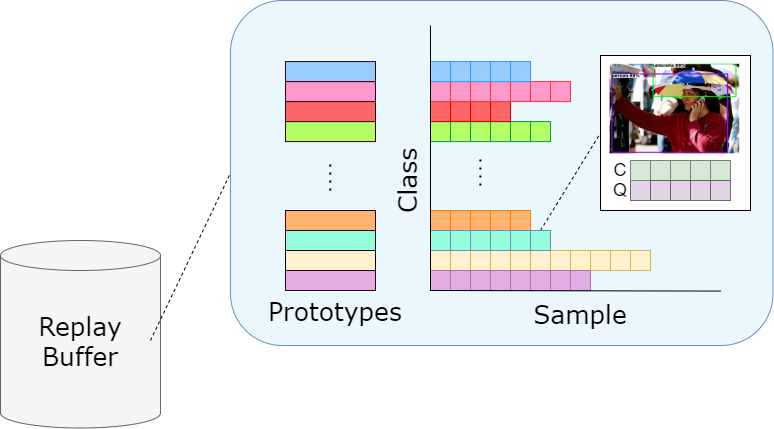} 
  \caption{An overview of the contents of our replay buffer. Each image is stored with its top-k class embeddings and their corresponding query embeddings. Note that classes are overlapping in terms of the samples they contain, as most samples contain instances of multiple classes.} 
  \label{fig:short-a} 
\end{figure}

All of our experiments used the OWL-ViT foundation model (Figure~\ref{fig:owlvit}), which is a CLIP model adapted to object detection by removing the final pooling layer of the vision transformer and attaching a lightweight class embedding and box prediction head. For each image, the class embedding head outputs a tensor of size $(T, E)$, where $T$ is the number of tokens processed by the vision transformer and $E$ is the class embedding size. The box prediction head outputs a tensor of size $(T, 4)$ \cite{minderer2022simple}. See Figure~\ref{fig:short-a} for a depiction of the contents of our replay buffer; note that the term "prototype" refers to the average class embedding for a given class. 

\subsection{Class Selection Primitives}

Class selection primitives are those that determine the classes to be sampled from for each new batch. Typically, this means producing a different probability distribution over classes for each image in the batch, then sampling a single class for each. In this work, we evaluate two class-selection primitives.

\textbf{Balanced:} Classes are sampled equally and deterministically. For example, if we are currently retrieving four samples from the replay buffer and the last sampled class was class 10, we would retrieve a sample from classes 11 through 14. If we were to retrieve four more samples for the next batch, we would retrieve a sample from classes 15 through 18.

\textbf{Similarity-Weighted \cite{saxena2022learning}:} Classes are sampled in proportion to their prototypes' similarities with new samples' embeddings (See Section~\ref{sec:fullswil} of the Supplementary Material for a full description of the algorithm). In general, we make two important modifications with respect to \cite{saxena2022learning}, since we adapt the original formulation to operate fully online (on the level of batches) and also use a hyperparameter to control the entropy of the similarity-weighted distribution over classes. 

\subsection{Sample Selection Primitives}

Sample selection primitives are those that produce distributions over samples in the replay buffer. In this work, we evaluate three of them. Note that our first two sample selection primitives assume a single class to be sampled from (Figure~\ref{fig:short-a}) while our third sample selection primitive does not make this assumption, only assuming access to a \textit{candidate set} of samples.

\textbf{Uniform:} Each sample in the selected class has an equal probability of being selected.

\textbf{Prototype-Weighted \cite{harun2023grasp}:} Each sample in the selected class has a probability in inverse proportion to its distance to its corresponding class prototype. We begin computing each sample's "score" by taking the minimum cosine distance between its top-$k$ class embeddings \textit{for that class} and the class prototype, where $k$ is set as the ground-truth number of instances for the class within the sample's image. We finalize each sample's score by raising its distance to a negative power, which is controlled by a hyperparameter. Thus, the full equation for computing a sample's score is:

\begin{equation}
\label{eq:proto}
    s_i = \min_{j \in k} \left(1 - \frac{\mathbf{e}_c \cdot \mathbf{e}_j}{\|\mathbf{e}_c\| \|\mathbf{e}_j\|}\right)^{-w}
\end{equation}
where $e_j$ is a class embedding from the sample, $e_c$ is the class prototype, and $w$ controls the entropy of the distribution of scores across samples.

We then convert each sample's score into a probability using:

\begin{equation}
    p_i = \frac{s_i}{\sum_{s_m \in N_c} s_m}
\end{equation}
where $N_c$ represents the number of samples containing an instance of class $c$.

\textbf{Adversarial Shapley Value \cite{shim2021online}:} Each sample in a selected \textit{candidate set} is scored according to the Adversarial Shapley Value (ASV), which is derived from the KNN Shapley Value (KNN-SV) \cite{jia2020efficient}. The KNN Shapley Value is an efficient approximation of the Shapley Value \cite{shapley1953value} used in data valuation and can be understood as an algorithm for attributing predictions made by a model for individual samples in an \textit{evaluation set} (\eg a validation set or, in this case, a batch of new data) to individual samples in a training set (which, in our case, is instead the \textit{candidate set} drawn from our replay buffer). When computing a set of KNN-SVs for a candidate set $D_c$ with $N_c$ images and an evaluation set $D_e$ with $N_e$ images, our goal is to compute a $(N_c, N_e)$ tensor of KNN-SVs, where a \textit{high} KNN-SV at index $(i,j)$ means that the model's prediction for the evaluation image $j$ is \textit{highly} attributable to candidate image $i$ being present in the training dataset. 

Proposed specifically for selective retrieval from a replay buffer, the \textit{Adversarial} Shapley Value is computed for each sample in the candidate set as:
\begin{equation}
    ASV(i) = \frac{w}{N_c} \sum_{j \in D_c} s_j(i) - \min_{k \in D_e} s_k(i)
    \label{eq:ASV}
\end{equation}
where $s_j(i)$ is the KNN-SV of sample $i$ with respect to sample $j$ and $w$ is a weighting term controlled by a hyperparameter. Note that the left term uses $D_c$ as the candidate set \textit{and} as the evaluation set, and is meant to convey how \textit{representative} sample $i$ is of the other candidates. The right term uses $D_c$ and $D_e$ in their usual roles and is subtracted so that it conveys how \textit{adversarial} sample $i$ is with respect to the evaluation set, which is a batch of new samples in our case. 
While \cite{shim2021online} found that averaging both terms performs best, we found averaging the left term and minimizing the right term outperform all other combinations. 

In introducing $w$, we find it beneficial to dynamically compute its value as opposed to fixing it:
\begin{equation}
    w = c * \frac{\|\min_{s \in S_r}s\|}{\|\min_{s \in S_l}s\|}
    \label{eq:ASVw}
\end{equation}
where $c$ is a hyperparameter, $S_r$ is the set of KNN-SVs computed for the right term of the ASV and $S_l$ is the set for the left term. 

See Section~\ref{sec:fullasv} of supplementary material for a full description of the algorithm, including the computation of the KNN-SV.

\subsection{Composing and Modifying Primitives}

Our main results (Table~\ref{tab:algo}, ~\ref{tab:main}) consider combinations of class and sample selection primitives. We refer to each composition of primitives using acronyms from the original papers. (\eg "ASER" refers to combining balanced class selection and ASV sample selection). Besides combining existing primitives (and adding hyperparameters), we extend a few primitives with respect to how they are combined and/or implemented:
\begin{enumerate}
    \item ASER-\textit{PC}, where the left term of the ASV is \textit{pre-computed} with the \textit{entire replay buffer} as the candidate and evaluation dataset. This allows us to scale the candidate set and potentially get a more accurate estimate of how representative each sample is of the replay buffer.
    \item \textit{A}-SW-GRASP, where we \textit{adaptively} determine whether to use GRASP (balanced class selection, prototype-weighted sample selection \cite{harun2023grasp}) or SWIL (similarity-weighted class selection, uniform sample selection \cite{saxena2022learning}) for each sample in the batch. We use GRASP when the normalized entropy of the distribution over classes produced by SWIL is too close to 1.0, where the specific threshold is determined by a hyperparameter.
\end{enumerate}

\subsection{Deduplication Schedules}

We refer to \textit{deduplication} as the process of reducing duplicate replays for a fixed time period, where after each period, all samples in the buffer are once again eligible for retrieval.

A \textit{deduplication schedule} is a function that determines the length of this time period; we study three of them in this work: ensuring that there are no duplicate replays occurring within each \textit{epoch}, within each \textit{dataset}, and within $\frac{1}{3}$ of the replay buffer that is replayed. In our case, the latter schedule corresponds to about the $\frac{1}{3}$-mark of training, on average.

\subsection{Loss Adaptivity}

New samples with low losses have smaller gradients and thus may be able to be learned without replay. Selectively limiting the impact of replay in this way could lead to more plasticity for new tasks, without incurring more forgetting. Here, we propose an algorithm for adaptively reducing the number of replayed samples and for reducing the magnitude of replayed samples' losses according to each new sample's loss magnitiude.

For each batch of new samples, we begin by computing each sample's loss. We then compute the number of samples that have loss values above a threshold:

\begin{equation}
    r = \left| \{ s \in B \mid \text{loss}(s) > l \} \right|
    \label{eq:la1}
\end{equation}

where $B$ is the current batch, $s$ is a sample from the current batch, and $l$ is a hyperparameter. We then use $r$ to determine how many samples to replay. After separately computing the average loss of samples in the new batch and the average loss of samples in the replay batch, we compute the final loss value for the batch as follows:

\begin{equation}
    L_T = L_B + \frac{r}{|B|}*L_R
    \label{eq:la2}
\end{equation}

where $L_T$ is the final loss value for the batch, $L_B$ is the average loss computed for the batch, and $L_B$ is the average loss computed for the replay batch.

\subsection{Replay Buffer Selection}

Before training, we must select samples from the pre-training dataset for our replay buffer. Although there are sophisticated methods, they require access to samples' gradients throughout training \cite{aljundi2019gradient, tiwari2022gcr}, which we do not have. We therefore propose a gradient-free algorithm for selectively constructing a replay buffer.

To satisfy the need for low-loss samples in the replay buffer \cite{hurtado2023memory}, we begin by removing samples that have loss values greater than a predefined magnitude:

\begin{equation}
    R = \{ s \in D_{pt} \mid \text{loss}(s) < t \}
    \label{eq:buffselect}
\end{equation}

where $R$ is the replay buffer, $s$ is an image in $D_{pt}$, and $t$ is a hyperparameter.

Then, instead of maximizing the diversity of gradients in the replay buffer, we sampled from $D_{pt}$ to ensure that each class had at least 50 images in $R$.
\section{Experiments}\label{sec:exps}

In this section, we detail the experimental setup used to evaluate all proposed methods. Our experiments begin with a full sweep over loss thresholds for constructing the replay buffer. We then present our main results, which represent a comprehensive and simultaneous sweep over deduplication schedules and hyperparameters for each selective retrieval algorithm. Next, we take a closer look at the effects of different deduplication schedules. Finally, we provide results for loss-adaptive retrieval.

\begin{table}
  \centering
  \begin{tabular}{@{}lcccc@{}}
    \toprule
    Replay Buffer & O365 & LVIS & LVIS rare & ODinW \\
    \midrule
    Non-selective & 20.6 & 19.8 & 15.5 & 31.6 \\
    Selective & \textbf{21.8} & \textbf{23.8} & \textbf{20.0} & \textbf{33.2} \\
    \bottomrule
  \end{tabular}
  \caption{Test mAP of best-performing replay buffer versus including all pre-training samples in the replay buffer. Uniform retrieval. The results are averaged across 3 dataset orderings and recorded at the end of the 15-dataset sequence.}
  \label{tab:buff}
\end{table}

\begin{table*}
  \centering
  \begin{tabular}{@{}l|cc|ccc|cc@{}}
    \toprule
    \multicolumn{1}{c}{} & \multicolumn{2}{c}{Class Selection} & \multicolumn{3}{c}{Sample Selection} & \multicolumn{2}{c}{Modifications} \\
    \cmidrule(lr){2-3} \cmidrule(lr){4-6} \cmidrule(lr){7-8}
     & Balanced & Similarity & Uniform & Prototype & ASER & ASER-PC & Adaptive \\
    \midrule
    Uniform &  &  &  &  &  &  &  \\ \hline
    Uniform balanced & \checkmark &  & \checkmark &  &  &  &  \\ \hline
    GRASP \cite{harun2023grasp} & \checkmark &  &  & \checkmark &  &  &  \\ \hline
    SWIL \cite{saxena2022learning} &  & \checkmark & \checkmark &  &  &  &  \\ \hline
    SW-GRASP &  & \checkmark &  & \checkmark &  &  &  \\ \hline
    A-SW-GRASP &  & \checkmark &  & \checkmark &  &  & \checkmark \\ \hline
    ASER \cite{shim2021online} & \checkmark &  & \checkmark &  & \checkmark &  &  \\ \hline
    ASER-PC & \checkmark &  & \checkmark &  &  & \checkmark &  \\ \hline
    SW-ASER-PC &  & \checkmark & \checkmark &  &  & \checkmark & \\
    \bottomrule
  \end{tabular}
  \caption{Retrieval algorithms evaluated in this work.}
  \label{tab:algo}
\end{table*}

\begin{table}
  \centering
  \begin{tabular}{@{}l|c|c|c|c@{}}
    \toprule
    Algorithm & O365 & LVIS & LVIS rare & ODinW \\
    \midrule
    \rowcolor{gray!25}
    No fine-tuning & 24.2 & 29.0 & 23.8 & 12.3 \\ 
    \rowcolor{gray!25}
    Separate models & - & - & - & 45.6 \\ 
    \rowcolor{gray!25}
    No replay & 0.0 & 0.3 & 0.2 & 11.1 \\ 
    \rowcolor{gray!25}
    Uniform & 21.8 & 23.8 & 20.0 & 33.2 \\ 
    \rowcolor{gray!25}
    Uniform balanced & 22.2 & 24.5 & 20.7 & 33.7 \\ 
    GRASP \cite{harun2023grasp} & \textbf{22.5} & \textbf{24.9} & 21.4 & 33.6 \\ 
    SWIL \cite{saxena2022learning} & \textbf{22.5} & \textbf{24.9} & \textbf{21.6} & 33.6 \\ 
    SW-GRASP & 22.1 & 24.4 & 20.9 & 33.5 \\ 
    A-SW-GRASP & 22.4 & 24.8 & 21.5 & 33.7 \\
    ASER \cite{shim2021online} & 22.2 & 24.3 & 20.5 & 33.6 \\ 
    ASER-PC & 22.4 & 24.6 & 20.8 & 33.7 \\
    SW-ASER-PC & \textbf{22.5} & 24.7 & 20.9 & \textbf{33.9} \\ 
    \bottomrule
  \end{tabular}
  \caption{Test mAP for each retrieval algorithm. The results are shown for the best deduplication algorithm (deduplication for each dataset) averaged across six ODinW dataset orderings. The results are recorded at the end of the 15-dataset sequence. Baselines are highlighted in gray.}
  \label{tab:main}
\end{table}

\textbf{Model:} We focus our experiments on the OWL-ViT L/14 open-vocabulary object detection model \cite{minderer2022simple} (Figure~\ref{fig:owlvit}). This model was CLIP \cite{radford2021learning} pre-trained on 6 billion image-text pairs, then adapted to object detection on the Visual Genome \cite{krishna2016visual} and Objects365 \cite{shao2019objects365} datasets, which total over 700,000 images. See Section~\ref{sec:owlvit}.

\textbf{Datasets:} For the continual learning phase, we sequentially fine-tune our model on a carefully selected 15-dataset subset of the Objects Detection in the Wild (ODinW) \cite{li2022elevater} benchmark. See Section~\ref{sec:datasets} of the supplementary material for more details on our datasets.

\textbf{Training:} Before fine-tuning the model \textit{sequentially} on each ODinW dataset, we \textit{independently} perform a hyperparameter sweep on each dataset over learning rates, epochs, and layerwise decay coefficients of learning rate. We also used a cosine learning rate scheduler with a single, linear warmup epoch. Optimization details and hardware can be found in Section~\ref{sec:trainhps} of the supplementary material.

\textbf{Evaluation:} Before and after the continual learning phase, we evaluate the model on the Objects365 dataset and the LVIS dataset. The LVIS dataset measures the open vocabulary performance of the model and, consistent with \cite{minderer2022simple}, we also evaluate the model on the "rare" class subset of LVIS, which includes classes for which the model has never seen bounding boxes. These three benchmarks allow us to evaluate the model on classes that are and are not in the replay buffer, which is necessary for this problem setting.

We follow the standard practice of measuring AP at IoU=.50:.05:.95, averaged across classes (termed mAP). We measure the performance of each algorithm according to mAP after training on all datasets, where "ODinW" indicates the average mAP across all 15 datasets.

\subsection{Replay Buffer Selection}

Table~\ref{tab:buff} compares the performances between continual learners using the entire pre-training dataset as the replay buffer and continual learners using the buffer produced by the \textit{best-performing} Objects365 and Visual Genome loss thresholds. It is clear that our simple, gradient-free buffer selection algorithm provides substantial performance gains relative to the non-selective baseline, especially for classes which are not represented in the replay buffer (LVIS and LVIS rare). 

It is also clear that our model shows a strong preference for low-loss samples in the replay buffer, as the best-performing hyperparameter set only included 50,000 images. The best replay buffer also entirely excluded images from Visual Genome, which can be explained by its extremely large concept space and high average loss per image; all VG images have more than 10x the loss of any Objects365 images in our final replay buffer.

Overall, the success of our buffer selection algorithm suggests that it is possible to replicate the success of more sophisticated gradient-dependent techniques with simple loss thresholding and class balancing.

\begin{table}
  \centering
  \begin{tabular}{@{}l|c|c|c|c@{}}
    \toprule
    Replay \% & O365 & LVIS & LVIS rare & ODinW \\
    \midrule
    \rowcolor{gray!25}
    100\% & \textbf{21.8} & \textbf{24.5} & \textbf{22.2} & 33.2 \\ 
    97\% & 20.9 & 23.4 & 19.9 & 33.3 \\
    85\% & 18.7 & 21.7 & 17.7 & 32.5 \\
    \bottomrule
  \end{tabular}
  \caption{Test mAP of each loss threshold when paired with SWIL, averaged across 3 random dataset orderings. The baseline is highlighted in gray.}
  \label{tab:la}
\end{table}

\begin{table}
  \centering
  \begin{tabular}{@{}l|c|c|c|c@{}}
    \toprule
    Deduplication & O365 & LVIS & LVIS rare & ODinW \\
    \midrule
    \rowcolor{gray!25}
    None & 21.8 & 23.9 & 20.2 & 32.5 \\ 
    Each epoch & 21.8 & 24.2 & 20.5 & 32.1 \\ 
    Each dataset & \textbf{22.1} & \textbf{24.5} & \textbf{21.1} & 32.9 \\ 
    1/3 of buffer & 21.6 & 23.7 & 20.0 & 32.7 \\
    \bottomrule
  \end{tabular}
  \caption{Test mAP of each deduplication algorithm when paired with SWIL, averaged across 3 random dataset orderings. The baseline is highlighted in gray.}
  \label{tab:dedup}
\end{table}

\begin{figure*}
    \centering
    \includegraphics[width=\linewidth]{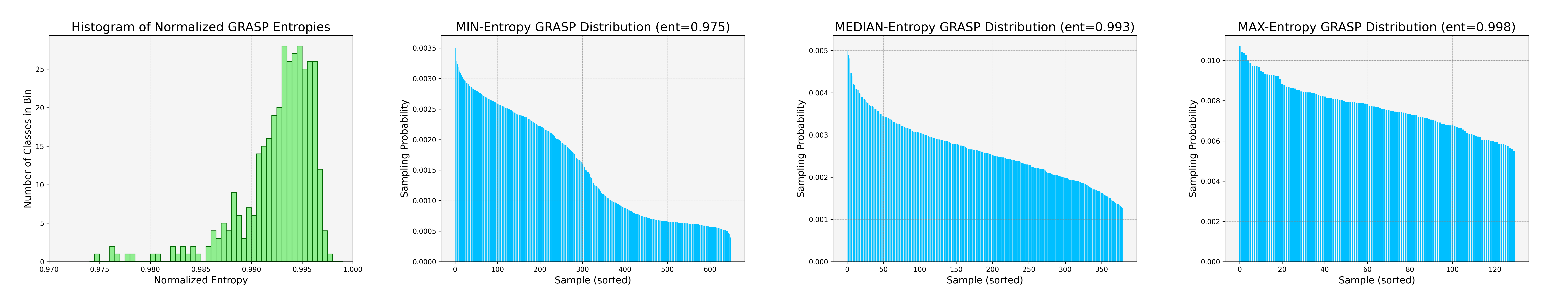}
    \captionsetup{width=\linewidth, justification=centering}
    \caption{Histogram of normalized entropies for distributions produced by GRASP, followed by the distributions with minimum, median, and maximum entropies. Recall that GRASP produces a distribution over samples within each class.}
    \label{fig:gdist}
\end{figure*}

\subsection{Comparing Retrieval Strategies}

After building the replay buffer, we perform a hyperparameter sweep for each algorithm (refer to Table~\ref{tab:algo} for a description of each algorithm), which includes a sweep over deduplication schedules. Table~\ref{tab:main} reports the test performance of each algorithm's best-performing hyperparameter sets.

We find that the two top-performing algorithms are also the most simple and, most surprisingly, that combining the two best algorithms results in a significant drop in performance by all measures. The failure of A-SW-GRASP particularly indicates that the best class selection primitive (similarity-weighted) and the best sample selection primitive (prototype-weighted) are fundamentally incompatible despite being completely orthogonal. A potential explanation for this incompatibility is that both SWIL and GRASP can range from weak to strong selectivity in their produced distributions (see Section~\ref{sec:shapes}) which may, on average, excessively reduce the randomness of the overall algorithm. This effect may be especially prevelant considering the fact that each algorithm operates under the assumption that there is a single class per image, while there may be many classes per image in our setting.

Another surprising result is that using the adversarial shapley value provides little to no benefits over uniform balanced retrieval and is even counterproductive in its original form ("ASER"). A potential explanation for this outcome is that ASER shows a strong preference for the retrieval of samples which have the smallest distances to samples in the current batch (see Section~\ref{sec:ASERbias}).

Overall, the failures of SW-GRASP and A-SW-GRASP along with the marginal benefits of ASER-PC and SW-ASER-PC suggest that the problem of selective retrieval involves a delicate tradeoff between selectivity and diversity.

\subsection{Preventing Duplicate Replays}

We now compare deduplication schedules. See Table~\ref{tab:dedup} for SWIL results. All algorithms besides uniform balanced show a preference for \textit{dataset-level} deduplication, with the worst-performing schedule consistently being the strictest.

This result reflects the fact that each new dataset represents a distinct continual learning problem (see Figure~\ref{fig:dssize}) which may call for maximum selectivity in the early epochs since that is when each new sample's gradients are largest and have the highest variance. This selectivity can then be reduced in later epochs in favor of greater coverage of the replay buffer.

\begin{figure}
    \centering
    \includegraphics[width=\linewidth]{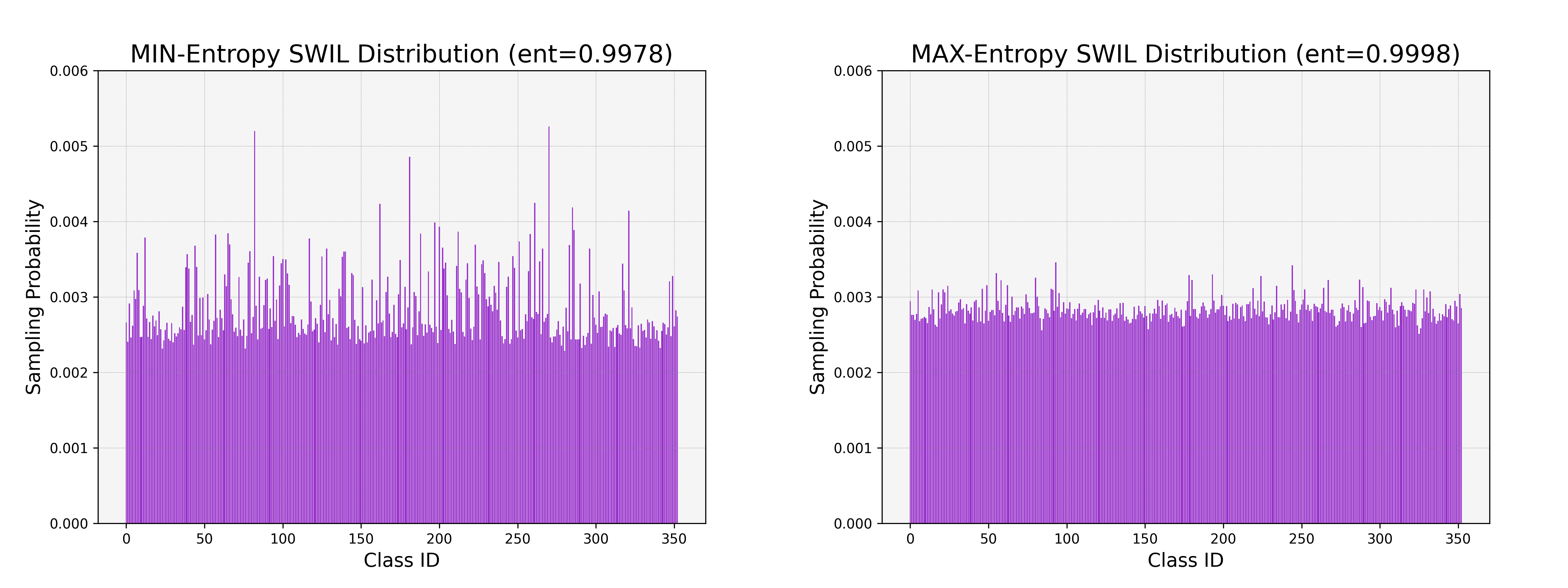}
    \captionsetup{width=\columnwidth, justification=centering}
    \caption{SWIL distributions with minimum (left) and maximum (right) entropies. Remember that SWIL produces a distribution over classes in the replay buffer.}
    \label{fig:sdist}
\end{figure}

\subsection{Loss Adaptivity}

Following our main experiments, we tested whether keeping a 1:1 ratio between new and replayed samples and maintaining the full impact of the replay loss, regardless of new samples' losses, is necessary. As seen in Table~\ref{tab:la}, the case where as little as 3\% of new samples leads to a performance decrease.

Paired with the "No Replay" row from Table~\ref{tab:main}, this result suggests that fine-tuning a \textit{foundation model} represents a setting where representations are \textit{more fragile} than in standard continual learning settings, as previous work has shown baseline training runs with no continual learning methods to catastrophically forget, but not to \textit{completely} destroy knowledge from previous tasks, especially when using proper training objectives. For example, \cite{asadi2023prototypesample} shows $30\%$ final accuracy on Split-Mini-ImageNet when naively fine-tuning the model with a SupCon \cite{khosla2021supervised} loss, while our model quickly drops to an mAP of around 0 when naively fine-tuning on new tasks.  
\section{Analyses}\label{sec:analyses}

In this section, we examine the distributions produced by SWIL and GRASP then search for characteristics of downstream datasets which are predictive of forgetting.

\subsection{The Shape of SWIL and GRASP}\label{sec:shapes}

The leftmost plot of Figure~\ref{fig:gdist} shows a histogram for the normalized entropies of the distributions produced by GRASP, while the three rightmost plots show the minimum, median, and maximum entropy distributions produced by GRASP as bar charts. Each of the three rightmost distributions are over samples within an O365 class and are measured before the continual learning phase.

The distribution of entropies has a long tail, and even the median-entropy distribution produced by GRASP is strongly selective (the highest-probablity sample has a ~5x higher probability than the lowest-probability sample). These plots therefore show that GRASP prevents forgetting by showing a strong preference towards prototypical samples throughout training. These plots also explain why omitting deduplication for GRASP leads to a substantial performance decrease.

Figure~\ref{fig:sdist} shows the minimum- and maximum-entropy distributions (over classes) produced by SWIL for a dataset. We took the average probability of selecting each O365 class \textit{for an entire continual learning dataset} as the SWIL distribution for that dataset.

While the minimum-entropy SWIL distribution is quite selective (probabilities range from ~0.0025 to 0.005), the maximum-entropy distribution is nearly uniform.

To determine whether lower-entropy (more selective) SWIL distributions predict SWIL's per-dataset performance increase/decrease relative to GRASP, we computed a correlation between each dataset's SWIL entropy and the difference between SWIL and GRASP's Objects365 mAPs after training on the dataset. This resulted in a correlation coefficient of $-0.08$ (less entropy correlates with better performance). 

This relationship suggests that SWIL's equivalence with GRASP (Table~\ref{tab:main}) can be attributed to SWIL effectively targeting the most vulnerable circuits within the network in the case of sufficient similarity between new samples and samples in the pre-training dataset; in other cases, SWIL is outperformed by GRASP.

Importantly, this insight leads to the conclusion that SWIL may scale quite well with the number of classes in the replay buffer, as it would then be more likely to produce lower-entropy distributions over classes in the replay buffer. Such an effect could lead SWIL to outperform GRASP in certain scenarios, although the inverse would likely hold as well.

See Section~\ref{sec:sbias} of supplementary material for analysis of the relationship between the distributions produced by SWIL and resulting class-specific forgetting.

\begin{figure}
    \centering
    \includegraphics[width=\linewidth]{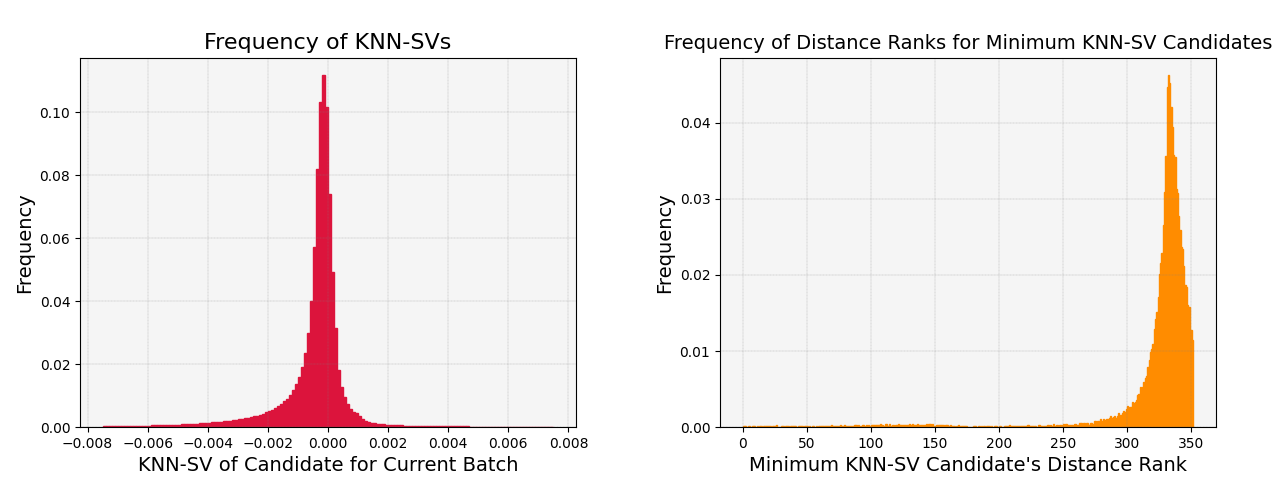}
    \captionsetup{width=\columnwidth, justification=centering}
    \caption{Histogram of KNN-SVs across an entire training sequence (left) and histogram of distance ranks for lowest-scoring (most likely to be selected) images in a candidate set across the same sequence (right). A higher distance rank means the image is closer to the evaluation image; each candidate set contained 352 images.}
    \label{fig:adist}
\end{figure}

\subsection{The Bias of ASER}\label{sec:ASERbias}

The left plot of Figure~\ref{fig:adist} contains a histogram for the KNN-SVs computed between candidate images and batch images for SW-ASER-PC (Table~\ref{tab:algo}). It is clear that the KNN-SV provides a strong signal, as the distribution is long-tailed.

The right plot of Figure~\ref{fig:adist} contains a histogram depicting the frequency of different distance rankings for the lowest-KNN-SV candidate images, where the "distance ranking" is the ranking of the candidate image in terms of its distance to an image in the current batch. Note that the image with the lowest KNN-SV for the batch is most likely to be retrieved, since the "adversarial" term of the ASV is negated (Equation~\ref{eq:ASV}).

From the right plot, it is clear that SW-ASER-PC is biased towards retrieving the from the ~25-closest candidates for each sample in the batch. This property leads to redundancy in retrieved samples, which may explain the lesser performance of SW-ASER-PC relative to SWIL and GRASP.

\begin{figure}[t]
  \centering
  \captionsetup{justification=centering}
  \includegraphics[width=\columnwidth]{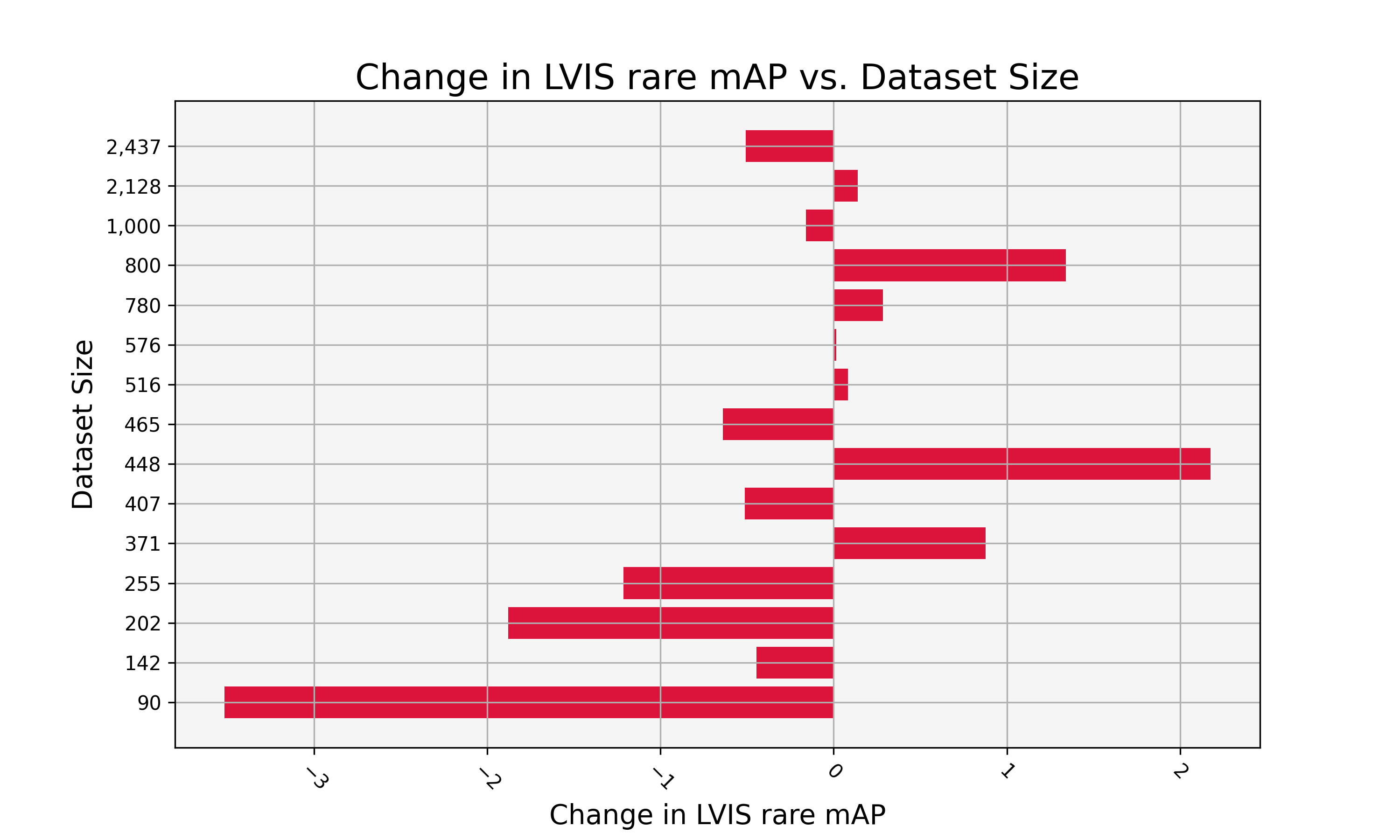}
  \caption{Change in LVIS rare mAP for each dataset, sorted by dataset size. Averaged across 3 SWIL training runs.}
  \label{fig:dssize}
\end{figure}

\subsection{Dataset-Dependent Forgetting Dynamics}

In our search for predictors of dataset-dependent forgetting, we correlated forgetting (mAP of pre-training datasets before and after encountering each downstream dataset) with characteristics including datasets' mAP before training, mAP improvement from training, position in the sequence of datasets, and size. 

The strongest predictor was datasets' \textit{size}, which is plotted in Figure~\ref{fig:dssize}. The rest of our metrics showed a weak correlation, if any. Note that pre-training mAP both \textit{improves} and \textit{degrades} throughout training. 

This result shows the destructive effect of reducing the time-to-recurrence of each datapoint, which follows from the intuition that continual learning calls for the prevention of model overfitting on unimportant aspects of new training data \cite{lesort2022continual}. The unreliability of the relationship shown in Figure~\ref{fig:dssize} also suggests more nuanced forgetting dynamics, which are likely dependent on properties of the model's pre-training distribution.

\section{Conclusion}

Selective retrieval is a key component of any replay algorithm. Existing methods can be organized into class- and sample-selective primitives; however, the best class- and sample-selective primitive cannot be combined to reach a new state of the art. This may be explained by the best selective retrieval algorithms being highly selective at times, leading to an overall insufficient amount of randomness across the learning trajectory.

Any algorithm used for selective retrieval should ensure that no duplicates are retrieved within a single continual learning task (dataset). This leads to retrieval being highly selective during the early epochs for each task (when losses on new samples are highest) and greater coverage of the replay buffer in later epochs.

When using replay, it is essential that the replay loss's magnitude is kept consistent, regardless of new samples' losses. This reflects the necessity of maintaining existing representations in the network, especially in the problem setting of continual learning \textit{at scale}, since previous representations cover many different classes and are required for forward transfer (optimal learning of new tasks).

Extensions to this work includes measures for reducing the computational overhead associated with replaying full inputs. This can be achieved by developing methods which replay old samples on the level of tokens, with an architecture similar to that of retrieval augmented generation \cite{lewis2021retrievalaugmented}. 

One key limitation of this work is that the replay buffer was not updated with samples from downstream tasks. Adding new samples to the replay buffer and ensuring they are replayed could potentially remove the performance gap between the ODinW performance upper bound and ODinW performance achieved when using replay.

Another key limitation of this work is that the continual learning phase only consisted of a sequence of 15 datasets. In practice, this sequence would need to be much longer for sequential learning to make sense, as all 15 datasets could be learned jointly. Some interesting properties of the problem setting also may only be revealed given a task sequence of sufficient length.

Based on the preceding analyses on dataset-dependent forgetting, a higher replay ratio for smaller downstream datasets should also be explored.

\section{Acknowledgments} 
This work is partially supported by the NSF EFRI BRAID grant \#2317706. 

{
    \small
    \bibliographystyle{ieeenat_fullname}
    \bibliography{main}
}

\clearpage
\setcounter{page}{1}
\maketitlesupplementary

\begin{figure}[t]
  \centering 
  \includegraphics[width=\columnwidth]{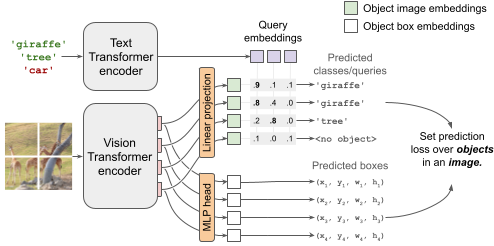} 
  \captionsetup{justification=centering}
  \caption{Depiction of the OWL-ViT model under study, taken from the original paper \cite{minderer2022simple}. Following CLIP pre-training, ViT token pooling is removed and light-weight object classification and localization heads are directly attached to the image encoder output tokens.} 
  \label{fig:owlvit} 
\end{figure}

\section{OWL-ViT}
\label{sec:owlvit}

Besides broad pre-training and 0.5B parameters, a desirable property of this model is that there are no task-specific parameters, which have complicated continual learning research in the past due to models experiencing abrupt representation drift upon the introduction of new, randomly initialized parameters \cite{mai2021supervised, caccia2022new, verwimp2021rehearsal, ramasesh2020anatomy}. 

\section{Full Similarity-Weighted Class Selection Primitive}
\label{sec:fullswil}
SWIL is executed in five main steps:
\begin{enumerate}
    \item Select the top-$k$ class embeddings for each image in the current batch; will be of size $(B, k, E)$ where $B$ is the batch size. Note that, here, top-$k$ means the $k$ class embeddings with the highest confidence score for \textit{any} query embedding.
    \item Compute each class embedding's similarity with each previous-class prototype; $(B, k, C_{rb})$, where $C_{rb}$ is the number of classes in the replay buffer
    \item Average or maximize similarities over dimension $k$ to get image-level similarities; $(B, C_{rb})$
    \item Transform these similarities into a probability distribution over previous classes; $(B, C_{rb})$
    \item Sample from each distribution to get a class ID for each sample in the batch.
\end{enumerate}

First, we select the top-$k$ class embeddings from each image in the current batch according to the confidence associated with each class embedding's predicted class. Assuming our model outputs a class probability tensor of shape $(\text{batch size}, \text{\# class embeddings}, \text{\# text queries})$, the indices of the top-$k$ class embeddings can be defined as the $(\text{batch size}, \text{k})$ tensor containing indices corresponding to the $k$ class embeddings which have the most positive inner product with \textit{any} of the text embeddings for the corresponding image (see Figure~\ref{fig:owlvit}). 

The next step is to compute the similarity between class embeddings of the current batch and a stored set of prototypical class embeddings, which includes one class embedding for each class learned before dataset $d_{\text{cl,i}}$. Assuming $C_\text{buff}$ classes are represented in the replay buffer, we are given a $(C_\text{buff}, \text{embedding dim})$ tensor of previous-class prototypes along with a  $(\text{batch size}, \text{k}, \text{embedding dim})$ tensor of current-batch class embeddings. We then compute a $(\text{batch size}, \text{k}, C_\text{buff})$ tensor of cosine distances, where cosine distance is defined as:

\begin{equation}
\label{eq:cosine_distance_ei_ej}
\text{dist}(\mathbf{e}_i, \mathbf{e}_j) = 1 - \frac{\mathbf{e}_i \cdot \mathbf{e}_j}{\|\mathbf{e}_i\| \|\mathbf{e}_j\|}
\end{equation}

To finalize these prototype similarities such that they can be interpreted on the level of \textit{images} as opposed to \textit{instances}, we minimize over dimension k to get a $(\text{batch size}, C_\text{buff})$ tensor of minimum prototype distances for each image.

The final step is to compute a weighted probability distribution over each class $C_\text{buff}$ for each image in the current batch, which will be proportional to the inverse of the image's prototype distance vector $\mathbf{d}_i$:

\begin{equation}
    \label{eq:swilprob}
    \mathbf{p}_{C_{\text{buff}}} = {\frac{1}{Z} \cdot \left(\mathbf{d}_i\right)^w}
\end{equation}

Which we then use to sample from each batch image's class distribution to determine which class to sample from for each image in the batch. 

\section{Full ASV Sample Selection Primitive}
\label{sec:fullasv}
Assuming a single discrete class label $y$ and a single class embedding per image, given evaluation point $(\mathbf{x}_j^e, \mathbf{y}_j^e) \in D_e$  we compute the KNN-SV using the following recursion: 
\begin{equation}
    s_j(\alpha_{N_c}) = \frac{\mathbbm{1}[y_{\alpha_{N_c}} = y_j^{ev}]}{N_c}
\end{equation}
\begin{equation}
\begin{gathered}
    \hspace{-12em} s_j(\alpha_m) = s_j(\alpha_{m+1}) + \\[1ex]
    \hspace{2em} \frac{\mathbbm{1}[y_{\alpha_m} = y_j^{ev}] - \mathbbm{1}[y_{\alpha_{m+1}} = y_j^{ev}]}{K} \frac{\min(K, m)}{m}
\end{gathered}
\end{equation}
where $\alpha_m$ is the index of candidate sample with the $m^{th}$-closest embedding to the evaluation point, $s_j(i)$ is the KNN-SV for pair $(i,j)$, $K$ is a pre-defined constant, and $\mathbbm{1}$ is the indicator function. Note that, in our implementation, we replace values for $y_{\alpha_m}$ and $y_j^{ev}$ with text embeddings and replace the indicator function with the \textit{cosine similarity} between embeddings. Similar to our SWIL implementation, we again use the top-k class embeddings as a drop-in replacement for the class embeddings used in previous work. This means our $(N_c, N_e)$ tensor of KNN-SVs instead becomes a $(N_c, k, N_e, k)$ tensor computed on the level of \textit{instances} in an all-to-all manner.

Given the above procedure for computing a $(N_c, k, N_e, k)$ tensor of KNN-SVs for a given candidate and evaluation set, we now go over the computation of the ASV, which was specifically designed for selective retrieval from a replay buffer \cite{shim2021online}. Given a candidate set $D_c$ containing a subset of the replay buffer and an evaluation set $D_e$ containing samples from the current batch, the ASV for the candidate sample at index $i$ is defined as:  
\begin{equation}
    ASV(i) = \frac{w}{N_c} \sum_{j \in D_c} s_j(i) - \min_{k \in D_e} s_k(i)
    \label{eq:ASV}
\end{equation}
where $w$ is a novel hyperparameter controlling the relative contributions of each term to the final score. Note that the left term of the ASV computes KNN-SVs with the candidate set \textit{also} acting as the evaluation set, while the right term computes KNN-SVs with the candidate set and evaluation set in their usual roles and is \textit{subtracted}. The left term therefore measures how \textit{representative} each candidate sample is of the candidate set while the right term measures how \textit{adversarial} the sample is for the current batch.

Before providing the two KNN-SV tensors to the ASV function, we maximize the KNN-SVs in the $(N_c, k, N_c, k)$ tensor (the left term of the ASV) over each k dimension to get a $(N_c, N_c)$ image-level tensor. For the $(N_c, k, N_e, k)$ tensor (right term), we instead minimize over the k dimensions. We found maximizing/minimizing to outperform averaging.  

To get the final ASV tensor, we apply Equation~\ref{eq:ASV} and linearly combine the tensors of the left term and right term after averaging and minimizing over each of their second dimensions, respectively; this leaves us with a 1-dimensional ASV tensor with $N_c$ entries. Finally, we take the top-n highest-scoring candidates as replay samples. Note that the original paper \cite{shim2021online} finds averaging both terms of the ASV to perform best, while we found averaging the left term and minimizing the right term to outperform all other combinations.

\section{Optimization and Hardware}
\label{sec:trainhps}

For superior regularization, we use Dark Experience Replay (DER++) \cite{buzzega2020dark}. DER++ offers substantial performance benefits over only using a classification and box-prediction loss on replayed samples thanks to its \textit{logit distillation} term. Note that we keep our replay buffer's logits and embeddings (including prototypes) fixed throughout the training sequence and that we do not add data from the continual learning sequence to the replay buffer.

All experiments were executed on a single DGX A100-640GB node; each training run took roughly 4 hours to complete. All selective retrieval algorithms incur negligible training overhead, except for ASER. Since ASER's overhead is controlled by the size of the candidate set and whether the left term of the ASV is pre-computed, we limit the size of the candidate set for each variant such that training-time overhead does not exceed 1\% of the training time of the "uniform balanced" algorithm.

\begin{table*}
  \centering
  \begin{tabular}{@{}l|c|c|c|c@{}}
    \toprule
    Dataset & \# Concepts & \# Train & \# Test & Pre-Trained OWL-ViT L/14 mAP \\
    \midrule
    Pothole & 1 & 465 & 67 & 23.0 \\ 
    WildfireSmoke & 1 & 516 & 74 & 24.4 \\ 
    ThermalCheetah & 2 & 90 & 14 & 18.7 \\ 
    ThermalDogsAndPeople & 2 & 142 & 20 & 49.9 \\ 
    BCCD & 3 & 255 & 36 & 3.6 \\
    ShellfishOpenImages & 4 & 407 & 58 & 19.9 \\
    EgoHands (specific)* & 4 & 1000 & 480 & 2.1 \\ 
    AerialMaritimeDrone(tiled) & 5 & 371 & 32 & 12.5 \\
    BrackishUnderwater* & 6 & 800 & 1468 & 2.5 \\
    Dice & 6 & 576 & 71 & 0.3 \\
    Aquarium & 7 & 448 & 63 & 21.6 \\
    ChessPieces & 13 & 202 & 29 & 3.5 \\
    AmericanSignLanguageLetters* & 26 & 780 & 72 & 0.6 \\
    Plantdoc & 30 & 2128 & 239 & 1.0 \\
    OxfordPets(breed) & 37 & 2437 & 345 & 0.7 \\
    \bottomrule
  \end{tabular}
  \captionsetup{justification=centering}
  \caption{ODinW datasets used in this work. * indicates that we use a balanced subset.}
  \label{tab:ds}
\end{table*}

\section{Datasets}
\label{sec:datasets}

We select our subset such that it solely consists of datasets which: have large enough test sets for low test-time variance, have high-quality, comprehensive annotations, and, most importantly, are performed poorly on by our pre-trained model (12.3 mAP average).

\begin{figure}
    \centering
    \includegraphics[width=\linewidth]{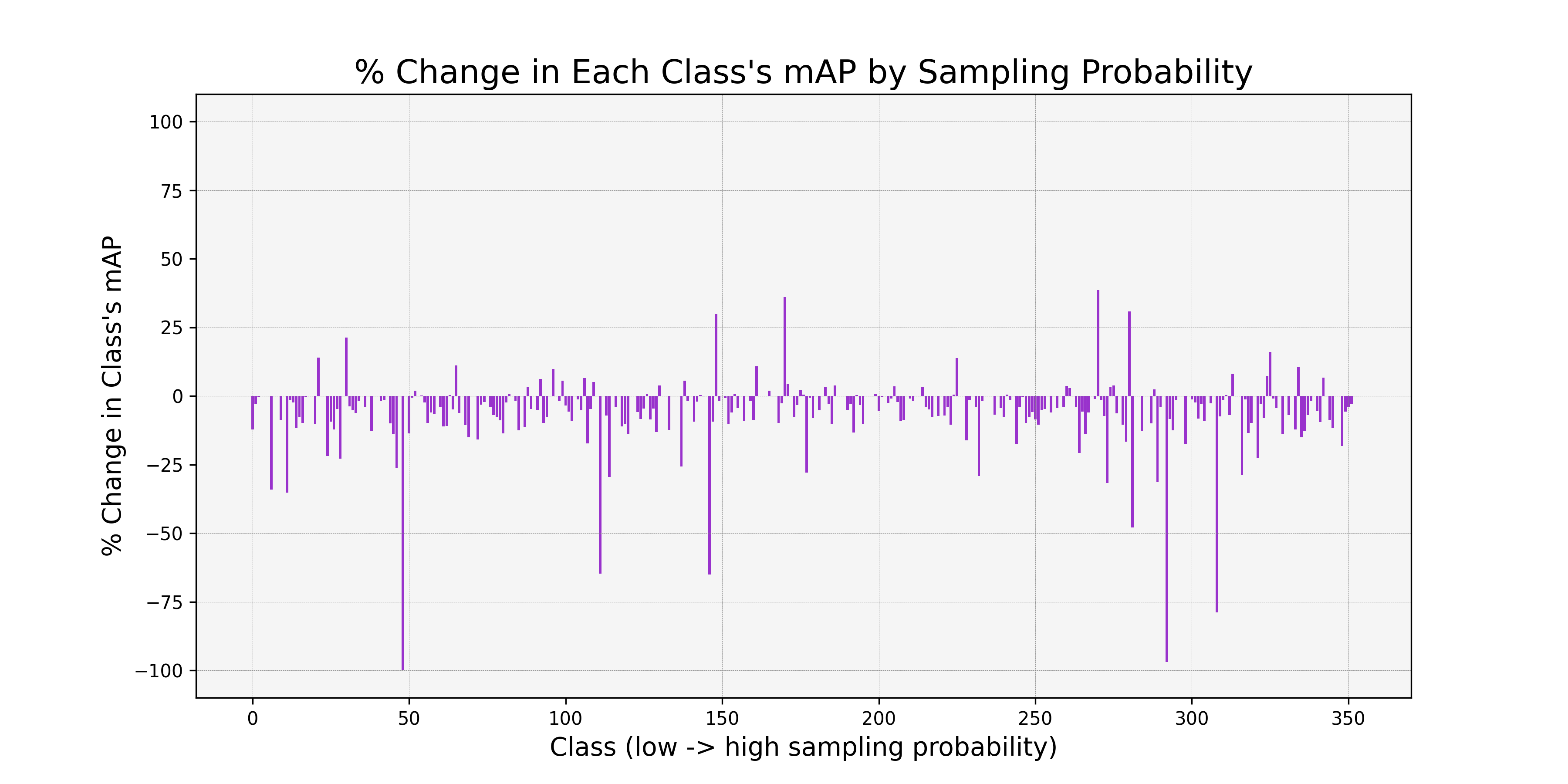}
    \captionsetup{width=\columnwidth, justification=centering}
    \caption{SWIL distributions with minimum and maximum entropies. Remember that SWIL produces a distribution over classes in the replay buffer.}
    \label{fig:sbias}
\end{figure}

\section{Is SWIL Biased?}
\label{sec:sbias}

It remains unclear whether SWIL's performance benefits when producing low-entropy distributions over classes is due to an imbalanced reduction in Objects365 classes' forgetting. Specifically, SWIL could be disproportionately improving performance on classes which it gives high probabilities to while ignoring others.

Figure~\ref{fig:sbias} shows the forgetting observed for each class when training on the dataset for which SWIL produced the lowest-entropy (most selective) distribution over classes. Class IDs (on the x-axis) are sorted in ascending order by their average sampling probabilities across the dataset. 

This plot shows that there is not a strong relationship between the forgetting of the model on specific classes and their average sampling probability across the dataset's training. This suggests that the performance benefits of SWIL are highly distributed across classes/subnetworks.

\section{Hyperparameters}
\label{sec:hps}

\begin{table}[b]
  \centering
  \begin{tabular}{@{}l|c|c@{}}
    \toprule
    Algorithm & Hyperparameter & Value \\
    \midrule
    DER$^{++}$ (replay loss) & $\alpha$ & 2.0 \\
    & $\beta$ & 1.0 \\
    All methods & k (for top-k tokens) & 8 \\
    Buffer Loss Thresholds & Objects365 & 0.15 \\
    & VG & 0.0 \\
    SWIL & w & 1.0 \\
    GRASP & w & 1.0 \\
    ASER & w & 0.15 \\
    & K (in recursive KNN-SV) & 20 \\
    & N candidates & 168 \\
    ASER-PC & N candidates & 352 \\
    \bottomrule
  \end{tabular}
  \caption{Hyperparameters used for each retrieval algorithm in this work. All were swept.}
  \label{tab:hps}
\end{table}

\begin{table*}
  \centering
  \begin{tabular}{@{}l|c|c@{}}
    \toprule
    Dataset & Hyperparameter & Value \\
    \midrule
    Pothole & Epochs & 5 \\ 
    & Linear Warmup Epochs & 1 \\
    & ViT LR & \(2.56 \times 10^{-4}\) \\
    & Text Encoder LR & \(2.56 \times 10^{-5}\) \\
    & LW-LRD Coeff & 0.99 \\
    WildfireSmoke & Epochs & 5 \\ 
    & Linear Warmup Epochs & 1 \\
    & ViT LR & \(1.4 \times 10^{-4}\) \\
    & Text Encoder LR & \(1.4 \times 10^{-5}\) \\
    & LW-LRD Coeff & 0.99 \\
    ThermalCheetah & Epochs & 7 \\ 
    & Linear Warmup Epochs & 2 \\
    & ViT LR & \(1.4 \times 10^{-4}\) \\
    & Text Encoder LR & \(1.4 \times 10^{-5}\) \\
    & LW-LRD Coeff & 0.99 \\
    ThermalDogsAndPeople & Epochs & 5 \\ 
    & Linear Warmup Epochs & 1 \\
    & ViT LR & \(1.4 \times 10^{-4}\) \\
    & Text Encoder LR & \(1.4 \times 10^{-5}\) \\
    & LW-LRD Coeff & 0.99 \\
    BCCD & Epochs & 7 \\ 
    & Linear Warmup Epochs & 2 \\
    & ViT LR & \(2.56 \times 10^{-4}\) \\
    & Text Encoder LR & \(2.56 \times 10^{-5}\) \\
    & LW-LRD Coeff & 0.99 \\
    ShellfishOpenImages & Epochs & 7 \\ 
    & Linear Warmup Epochs & 2 \\
    & ViT LR & \(2.56 \times 10^{-4}\) \\
    & Text Encoder LR & \(2.56 \times 10^{-5}\) \\
    & LW-LRD Coeff & 0.99 \\
    \bottomrule
  \end{tabular}
  \caption{Hyperparameters used for each dataset in this work. "LR" refers to the learning rate. "LW-LRD Coeff" refers to the LayerWise LR Decay Coefficient, which scales the LR at each layer (starting from the classifier and applied to the vision and text towers independently). All datasets used cosine LR decay with a linear warmup. All HPs were swept. * indicates that the dataset was subsetted (see Datasets table).}
  \label{tab:hps}
\end{table*}

\begin{table*}
  \centering
  \begin{tabular}{@{}l|c|c@{}}
    \toprule
    Dataset & Hyperparameter & Value \\
    \midrule
    EgoHands (specific)* & Epochs & 5 \\ 
    & Linear Warmup Epochs & 1 \\
    & ViT LR & \(1.4 \times 10^{-4}\) \\
    & Text Encoder LR & \(1.4 \times 10^{-5}\) \\
    & LW-LRD Coeff & 0.99 \\
    AerialMaritimeDrone(tiled) & Epochs & 7 \\ 
    & Linear Warmup Epochs & 2 \\
    & ViT LR & \(1.4 \times 10^{-4}\) \\
    & Text Encoder LR & \(1.4 \times 10^{-5}\) \\
    & LW-LRD Coeff & 0.98 \\
    BrackishUnderwater & Epochs & 7 \\ 
    & Linear Warmup Epochs & 2 \\
    & ViT LR & \(2.56 \times 10^{-4}\) \\
    & Text Encoder LR & \(2.56 \times 10^{-5}\) \\
    & LW-LRD Coeff & 0.99 \\
    Dice & Epochs & 7 \\ 
    & Linear Warmup Epochs & 2 \\
    & ViT LR & \(2.56 \times 10^{-4}\) \\
    & Text Encoder LR & \(2.56 \times 10^{-5}\) \\
    & LW-LRD Coeff & 0.99 \\
    Aquarium & Epochs & 5 \\ 
    & Linear Warmup Epochs & 1 \\
    & ViT LR & \(1.4 \times 10^{-4}\) \\
    & Text Encoder LR & \(1.4 \times 10^{-5}\) \\
    & LW-LRD Coeff & 0.99 \\
    ChessPieces & Epochs & 7 \\ 
    & Linear Warmup Epochs & 2 \\
    & ViT LR & \(6.4 \times 10^{-4}\) \\
    & Text Encoder LR & \(6.4 \times 10^{-5}\) \\
    & LW-LRD Coeff & 0.94 \\
    AmericanSignLanguageLetters* & Epochs & 7 \\ 
    & Linear Warmup Epochs & 2 \\
    & ViT LR & \(2.56 \times 10^{-4}\) \\
    & Text Encoder LR & \(2.56 \times 10^{-5}\) \\
    & LW-LRD Coeff & 0.96 \\
    Plantdoc & Epochs & 5 \\ 
    & Linear Warmup Epochs & 1 \\
    & ViT LR & \(2.56 \times 10^{-4}\) \\
    & Text Encoder LR & \(2.56 \times 10^{-5}\) \\
    & LW-LRD Coeff & 0.99 \\
    OxfordPets(breed) & Epochs & 5 \\ 
    & Linear Warmup Epochs & 1 \\
    & ViT LR & \(2.56 \times 10^{-4}\) \\
    & Text Encoder LR & \(2.56 \times 10^{-5}\) \\
    & LW-LRD Coeff & 0.99 \\
    \bottomrule
  \end{tabular}
  \caption{Continued}
  \label{tab:hps}
\end{table*}

\end{document}